\documentclass[cameraready]{Interspeech}

\title{Developing a Speech Language Model for Basque}

\author[affiliation={1},orcid=0009-0007-5354-684X, correspondingauthor]{Andoni}{Sudupe}
\author[affiliation={1},orcid=0000-0003-4447-7575]{Inma}{Hernáez Rioja}
\author[affiliation={1},orcid=0000-0003-0838-8516]{Ander}{Arriandiaga}
\author[affiliation={1},orcid=0000-0002-5013-1442]{Christoforos}{Souganidis}
\author[affiliation={1},orcid=0000-0001-7282-2765]{Ibon}{Saratxaga}
\author[affiliation={1},orcid=0000-0003-3804-4984]{Eva}{Navas}

\address{
    $^1$ HiTZ Basque Center for Language Technology - Aholab Signal Processing Laboratory, University of the Basque Country UPV/EHU, Spain 
}

\email{andoni.sudupe@ehu.eus, inma.hernaez@ehu.eus, ander.arriandiaga@ehu.eus, christoforos.souganidis@ehu.eus, ibon.saratxaga@ehu.eus, eva.navas@ehu.eus}


\usepackage{comment}
\usepackage{todonotes}
\usepackage{hyperref}
\usepackage{multirow}

\begin{document}

\maketitle

\begin{abstract}
Although the recent development of LLMs has paved the way for multi-modal interaction, minority languages face difficulties keeping pace. This work presents the first open Speech Language Model for Basque, designed for low-latency, end-to-end spoken interaction. Based on the LLaMA-Omni architecture, the system integrates a Latxa model as its backbone, transferring its Basque capabilities to the spoken modality by fine-tuning several modules. To address data scarcity, the English VoiceAssistant-400K corpus was translated into Basque, and the audio was synthesized using a Basque TTS. Evaluation results confirm the system's viability, achieving 3.86/5 in content quality, a 9.94\% WER for spoken intelligibility, and a 3.07 UTMOS for audio quality. This work demonstrates that advanced spoken interaction systems can be developed using modular architectures and synthetic data even in low-resource languages.

\end{abstract}
\section{Introduction}

\begin{figure*}
    \centering
    \includegraphics[width=1\linewidth]{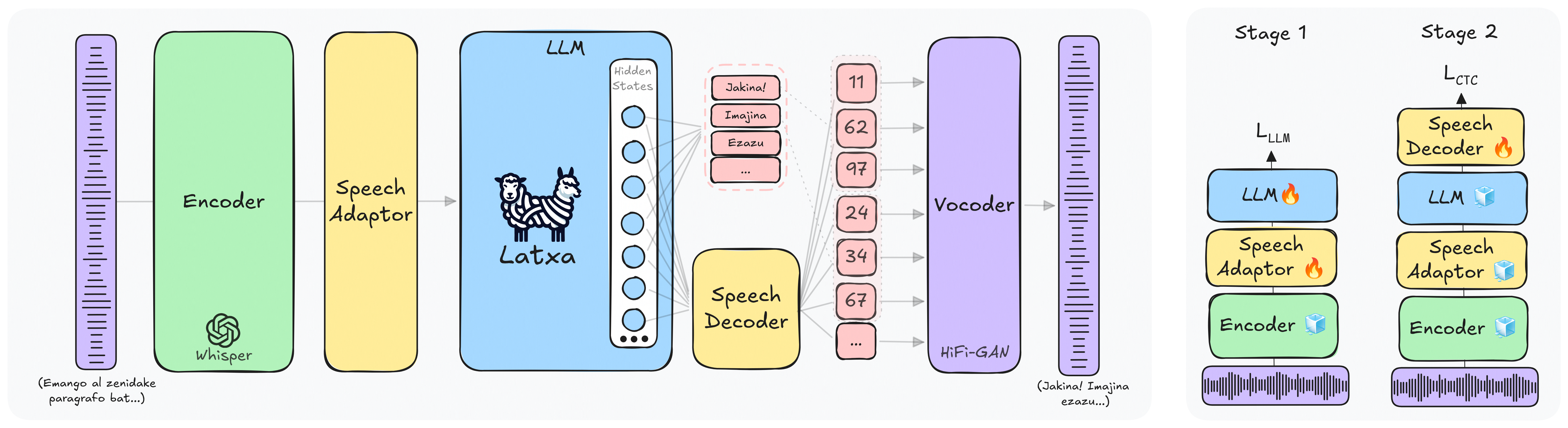}
    \caption{\textbf{Left}: Model architecture of our SLM, based on LLaMA-Omni. \textbf{Right}: Illustration of the two-stage training
strategy.}
    \label{fig:latxa-omni}
\end{figure*}

Recent developments in Large Language Models (LLMs) have paved the way for multi-modal interaction, combining textual data with other information modalities such as audio via Speech Language Models (SLMs). These systems bypass the traditional cascaded pipelines (ASR+LLM+TTS) to enable end-to-end spoken interactions. Processing speech natively mitigates the inherent limitations these cascaded pipelines have, like semantic information loss, error accumulation, and high response latency. Due to these capabilities, the most advanced proprietary systems have adopted natively multi-modal architectures capable of processing and generating speech. 

However, current SLMs are proprietary and primarily trained with English and Chinese resources, and therefore, minority languages face significant challenges in staying at the forefront of these advancements. While a performance gap already existed in text processing, introducing the speech modality worsens this disparity, as audio must inherently belong to the target language. Despite recent efforts in open-weight models like LLaMA-Omni~\cite{fang2025llamaomniseamlessspeechinteraction}, LLaMA-Omni2~\cite{fang2025llamaomni2llmbasedrealtimespoken} and Spirit LM~\cite{nguyen2024spiritlminterleavedspoken}, these systems remain almost exclusively focused on English and Chinese. 

Developing an SLM requires addressing critical design decisions and data scarcity, many of which have not been properly explored for low-resource languages. This paper investigates the development, training recipes, and evaluation methods required to build an open SLM for Basque. Following LLaMA-Omni, we use a Whisper model~\cite{radford2022robustspeechrecognitionlargescale} as an encoder and a Unit HiFi-GAN~\cite{polyak2021speechresynthesisdiscretedisentangled} vocoder trained from HuBERT~\cite{hsu2021hubertselfsupervisedspeechrepresentation} tokens to generate the output speech. Although centered on a single language (Basque), our study likely generalizes to other similarly resourced languages that struggle to keep pace with the multi-modal technological wave. 

To address the severe scarcity of Basque speech instruction data, we create the first multi-modal spoken-instruction datasets for Basque using a synthetic methodology. Relying on open resources, we translated the English VoiceAssistant-400K~\cite{xie2024miniomnilanguagemodelshear} corpus into Basque using the Latxa-3.1-70B-Instruct\footnote{\href{https://huggingface.co/HiTZ/Latxa-Llama-3.1-70B-Instruct}{Link to Latxa-3.1-70B-Instruct}}~\cite{sainz2025instructinglargelanguagemodels} model to ensure cultural and syntactic adaptation. Subsequently, the corresponding audio was synthesized using a Basque TTS system using different voices. 

With our newly generated dataset, we can adapt the Basque-instructed Latxa-3.1-8B-Instruct\footnote{\href{https://huggingface.co/HiTZ/Latxa-Llama-3.1-8B-Instruct}{Link to Latxa-3.1-8B-Instruct}}~\cite{sainz2025instructinglargelanguagemodels} model for multi-modal spoken tasks. In our experiments, we find that:  
\begin{itemize}
 \item Synthetic data is sufficient to build a robust SLM for a low-resource language: By translating existing English corpora and synthesizing the audio, we successfully trained a model that achieves a score of 3.86 out of 5 for content quality, based on LLM-as-a-Judge evaluation. 
 
 \item Modular architectures effectively transfer linguistic capabilities to the spoken modality: Our approach successfully bypasses the need for massive multi-modal pre-training. The resulting system achieves a normalized Word Error Rate (WER) of 9.94\% for Basque words, comparable to the intelligibility accuracy achieved by the original LLaMA-Omni model in English.
    
\end{itemize}

The remainder of the paper is organized as follows: Section \ref{section:2} details the system architecture of our modular Basque SLM. Section \ref{section:3} describes the data curation and synthetic generation pipeline used to construct the multi-modal corpora. Section \ref{section:4} outlines the multi-stage training methodology. Section \ref{section:5} presents the evaluation framework alongside the performance results, and Section \ref{section:6} summarizes our conclusions and discusses avenues for future work.

\section{System Architecture}\label{section:2}

In this work, we follow a modular design for the SLM inspired by LLaMA-Omni~\cite{fang2025llamaomniseamlessspeechinteraction}. This design choice is motivated by two main factors: (1) LLaMA-Omni is built upon the \textit{LLaMA-3.1-8B-Instruct}~\cite{dubey2024llama} backbone, for which a Basque-adapted version is already available; and (2) the use of pre-trained modules significantly reduces computational requirements during training, making it highly suitable for a low-resource language like Basque.

This modular approach involves attaching pre-trained modules to an already trained Large Language Model (LLM) to enable speech functionalities without requiring extensive architectural changes. This architecture is illustrated in Figure \ref{fig:latxa-omni}. First, an encoder processes the input speech into continuous representations. To enable the LLM to understand these extracted features, a trainable speech adaptor maps the speech representations directly into the LLM's embedding space. This mapping is achieved by first downsampling the sequence to reduce the temporal length. Then, it projects the representations through a two-layer Feed-Forward Network (FFN) with a ReLU activation function to match the LLM's embedding dimension. Once encoded, these representations are integrated into the LLM's prompt, allowing the model to produce the corresponding response.

LLaMA-Omni distinguishes itself from other models through its unique approach to processing outputs. Traditionally, decoding responses into speech requires intermediate text generation. In this architecture, however, the LLM's hidden states are decoded into speech and text in parallel, bypassing the dependency on intermediate text and enabling a very low response latency. The Speech Decoder takes the LLM's hidden states (captured right before text generation) and first upsamples them to match the temporal sequence length of the audio. Then a non-autoregressive (NAR) Transformer predicts discrete tokens. These tokens correspond to pre-trained clusters of speech representations extracted by an encoder, which allows a vocoder to synthesize speech directly from them.

During inference, to enable very low latency streaming synthesis, tokens are processed in chunks. When the number of generated tokens reaches a pre-defined chunk size $\Omega$, this segment is sent to the vocoder to synthesize a speech segment, which is then immediately played to the user. 

Following this architecture, we instantiate our system with Basque-specific components. The backbone of our system is Latxa-3.1-8B-Instruct, an LLM fine-tuned specifically for Basque based on the LLaMA-3.1 architecture. Leveraging this model ensures that the system maintains high syntactic, morphological, and semantic comprehension of the Basque language during spoken interactions. To encode the input speech, we employ the \textit{Whisper-large-v3-eu}~\cite{dezuazo2025whisperlmimprovingasrmodels} model, which has been adapted for Basque Automatic Speech Recognition (ASR). Finally, the discrete tokens that the Speech Decoder outputs are derived from a pre-trained mHuBERT-147~\cite{boito2024mhubert147compactmultilingualhubert} model. This encoder was fine-tuned for ASR (as described in sections \ref{subsection:hubert_dataset} and \ref{subsection:encoder_training}), and then grouped into 1000 clusters using $K$-Means. A Unit HiFi-GAN vocoder, trained specifically on Basque speech datasets in a multi-speaker configuration, generates continuous high-fidelity audio waveforms from these discrete units. 

\section{Data}\label{section:3}

As described in the design section, the SLM model needs different modules; a Whisper encoder, a HiFi-GAN vocoder, an encoder to extract discrete tokens (which is a HuBERT model) and an LLM. The Basque variant of the Whisper model and the LLM are available so we only need to train (1) the HuBERT to extract the discrete tokens, (2) the vocoder and the (3) end-to-end SLM. For this work we define 3 types of datasets; (1) a dataset with (audio, transcription) pairs for the HuBERT encoder, (2) a dataset with (discrete\_tokens, audio) pairs for the vocoder and (3) a dataset with (audio\_input, text\_output, discrete\_unit\_output) triplet data for the SLM.

\subsection{HuBERT training Dataset} \label{subsection:hubert_dataset}

There are two main methods to train the HuBERT encoder for Basque: conducting a continuous pretraining using HuBERT's Self Supervised framework or fine-tuning for a specific task. We opted for the latter approach, this decision is discussed in Section \ref{subsection:encoder_training}. To fine-tune a model for Automatic Speech Recognition (ASR), audio data and synchronized transcriptions are required. For this purpose, we used the \textit{composite-eu}\footnote{\href{https://huggingface.co/datasets/HiTZ/composite_corpus_eu_v2.1}{Link to composite-eu dataset}} dataset, which comprises three public datasets for Basque: CommonVoice ~\cite{ardila2020commonvoicemassivelymultilingualspeech}, BasqueParliament ~\cite{app14051951} and OpenSLR~\cite{kjartansson-etal-2020-open}. These collections were curated and combined to yield 600 hours of transcribed Basque audios.

\subsection{Vocoder Dataset}

Originally, HiFi-GAN vocoders use Mel Spectrograms to infer speech, but some work has been done to instead use discrete tokens extracted from an encoder~\cite{polyak2021speechresynthesisdiscretedisentangled}. To train the Unit HiFi-GAN vocoder, pairs of discrete units and audio are required, and for this purpose a Basque TTS dataset\footnote{\href{https://zenodo.org/records/17952596}{Link to TTS corpus}} was used. It contains 2 speakers: Maider (female) and Antton (male). Each speaker has 13,500 recorded utterances, with an average duration of 4 seconds per audio, totaling around 16 hours of speech per speaker.

\subsection{SLM Dataset}

Training a Speech Language Model requires data in a specific format. This training requires triplet data (audio\_input, text\_output, discrete\_unit\_output) structured in a speech-like or spoken-dialogue structure. This format is highly uncommon and no such public dataset is available for Basque.

To address this scarcity of speech instruction data in the Basque language, we built a synthetic workflow inspired by the construction of multi-modal datasets like VoiceAssistant-400K~\cite{xie2024miniomnilanguagemodelshear} or InstructS2S-200K~\cite{fang2025llamaomniseamlessspeechinteraction}. 

First, we extracted high-quality text instruction sets by translating the VoiceAssistant-400K dataset, which already follows the speech-like structure. For translation, we used the Latxa-3.1-70B-Instruct~\cite{sainz2025instructinglargelanguagemodels} model prompted with a strict rubric. This method removes the literal phrasing and parsing errors that conventional Machine Translation (MT) models often introduce.

Once the translation is completed, we used a VITS (Variational Inference with adversarial learning for end-to-end Text-to-Speech)~\cite{kim2021conditionalvariationalautoencoderadversarial} system to synthesise the utterances. The questions were synthesized using 14 synthetic voices distributed evenly by gender to ensure speaker diversity ~\cite{Hierro_Muga_Saratxaga_Couceiro_Hernaez_Rioja_Arriandiaga_Laresgoiti_Garcia_Romillo_2025}. For the target responses, we used the same speaker used to train the vocoder. 

Finally, for the second stage of the training, we extracted the discrete tokens from the target response audios. For the extraction we used the finetuned HuBERT for Basque (see section \ref{subsection:encoder_training}). To convert the response speech into discrete units, we first passed them through the fine-tuned HuBERT and selected the 9th hidden layer~\cite{polyak2021speechresynthesisdiscretedisentangled}. Then we applied a $K$-Means clustering algorithm optimized for $k=1000$ centroids~\cite{polyak2021speechresynthesisdiscretedisentangled}. The same HuBERT encoder and $K$-Means model must be used for the SLM and the vocoder, to avoid a mismatch in discrete tokens.

With this pipeline we successfully built a Basque spoken instruction dataset with approximately 450,000 questions with the corresponding answers, yielding around 550 hours of questions and over 3300 hours of answers, setting a viable baseline for end-to-end multi-modal alignment.

\section{Training}\label{section:4}

To develop our Basque Speech Language Model (SLM), the overall process first involves adapting the individual components; fine-tuning the speech encoder to extract discrete tokens and training the vocoder to synthesize acoustic waveforms. Once these components are prepared, the complete SLM undergoes a two-stage training process. The following subsections detail the specific training procedures for the encoder, the vocoder, and the end-to-end SLM architecture.

\subsection{HuBERT Training}\label{subsection:encoder_training}

To adapt the encoder for extracting Basque discrete tokens, we fine-tuned the model for an Automatic Speech Recognition (ASR) task. While continuous pre-training with new language data is typically the recommended approach, it requires a massive amount of data~\cite{hsu2021hubertselfsupervisedspeechrepresentation} that is currently unavailable for Basque. However, there is sufficient transcribed data to fine-tune the model for ASR.

As a starting point we selected the mHuBERT-147~\cite{boito2024mhubert147compactmultilingualhubert} checkpoint. This model was trained with 147 languages, including Basque, and has demonstrated better performance than much larger models across various tasks. 

The final encoder was trained using this configuration for 100 epochs with a batch size of 64. The training was conducted on a single NVIDIA H100 80GB GPU using the Hugging Face Transformers~\cite{wolf2020huggingfacestransformersstateoftheartnatural} library, taking approximately 400 hours to finish.
    
\subsection{Vocoder Training}

The selected vocoder, the Unit HiFi-GAN model, synthesizes waveforms directly from discrete representations. The training was implemented using the SpeechBrain~\cite{ravanelli2021speechbraingeneralpurposespeechtoolkit} framework. Based on early experimentation comparing single-speaker and multi-speaker configurations, we adopted a multi-speaker strategy, which demonstrated superior acoustic quality and stability. To condition the vocoder on different voices, speaker embeddings were extracted using a pre-trained \textit{ECAPA-TDNN}~\cite{Desplanques2020} model\footnote{\href{https://huggingface.co/speechbrain/spkrec-ecapa-voxceleb-mel-spec}{Link to the ECAPA-TDNN model}}. 

The final vocoder was trained on the two target voices (Maider and Antton), as previous tests showed a 2-speaker setup effectively captures necessary paralinguistic features without the complexity of broader configurations. It was trained for 1000 epochs with a batch size of 128. This process was executed on a single NVIDIA H100 80GB GPU and required roughly 10 hours to complete.

\subsection{SLM Training}

As shown in Figure \ref{fig:latxa-omni}, the training of the SLM is conducted in two distinct stages: (1) Speech Understanding and Text Generation and (2) Speech Response Generation. 

\textbf{Stage 1}
In the first stage, the primary objective is to teach the LLM the ability to understand spoken instructions and generate coherent text responses. During this phase, the speech encoder is kept completely frozen; only the speech adaptor and the LLM are trained. We train the LLM to output speech-like responses, prioritizing short and concise responses. The training is performed using a standard cross-entropy loss for next-token prediction.  

\textbf{Stage 2}
Once the model can understand speech and generate text, the second stage focuses on generating the spoken response in parallel. In this phase, the speech encoder, the speech adaptor, and the LLM are all frozen, and only the non-autoregressive streaming speech decoder is trained. Taking the hidden states of the LLM as input, the decoder learns to predict the sequence of discrete speech units. This alignment is learned asynchronously using a Connectionist Temporal Classification (CTC) loss, which manages the variable sequence lengths without requiring explicit text-to-speech alignment in the training data. 

For training we adapted the original codebase for LLaMA-Omni by implementing the missing training code and the dataset generation scripts. All the material and code used for this project is publicly available at 
\ifcameraready
    \href{https://huggingface.co/collections/Ansu/latxa-omni}{HuggingFace} and \href{https://github.com/ansuehu/Latxa-Omni}{Github},
\else
HuggingFace and Github,
\fi
respectively. 

For both training stages we followed the recommended hyperparameters by LLaMA-Omni. Each stage was executed for 3 epochs using a configuration of 4 NVIDIA A100 80GB GPUs, taking less than 100 hours in total. 

\section{Evaluation and Results}\label{section:5}

\begin{table*}[ht]
    \centering
    \begin{tabular}{|l|c|c|c|c|}
        \hline
        \multirow{2}{*}{\textbf{Model}} & \textbf{LLM} & \multirow{2}{*}{\textbf{WER (\%)}$\downarrow$}  & \multirow{2}{*}{\textbf{UTMOS}$\uparrow$} & \textbf{Streaming} \\
        & \textbf{Score (1-5)} $\uparrow$ & & & \textbf{Latency (ms)} $\downarrow$ \\
        \hline
        SpeechGPT (Eng) & 2.98 & 45.00 & 3.89 & 5587 \\
        SALMONN + Orca (Eng)& 3.44 & 3.78 & 3.82 & - \\
        Qwen2-Audio + Orca (Eng)& 3.47 & 6.77 & 3.62 & - \\
        LLaMA-Omni (Eng)& 3.99 & 10.82 & 3.92 & \textbf{346}\\
        LLaMA-Omni2-7B (Eng) & \textbf{4.28} & \textbf{3.26} & \textbf{4.22} & 582\\
        \hline
        \textbf{Ours (Eus)} & 3.86 & 9.94 & 3.07 & 370\\
        \hline
    \end{tabular}
    \caption{Comparison of our Basque (Eus) model with English (Eng) SLMs. English values are as reported in~\cite{fang2025llamaomniseamlessspeechinteraction} and \cite{fang2025llamaomni2llmbasedrealtimespoken}, using different judges/ASR systems; WER on synthesized speech also depends on TTS and ASR quality, so cross-language figures are indicative, not strictly comparable.}
    \label{tab:comparison_sota}
\end{table*}

To evaluate the performance of our architecture, we divided the evaluation into two main blocks: the individual components (the HuBERT discrete unit encoder and the Unit HiFi-GAN vocoder), and the complete end-to-end dialogue system.

\subsection{Modules}

\textbf{Discrete Unit Encoder (HuBERT):} Even though the HuBERT model is used to extract discrete units, it was fine-tuned on an Automatic Speech Recognition (ASR) task. To evaluate its learned representations we can use the standard metrics for ASR, Word Error Rate (WER) and Character Error Rate (CER). After training the model for 100 epochs, it achieved a 4.90\% WER on the test split of the \textit{composite-eu} dataset. Overall the encoder representations capture the phonetic properties of Basque effectively, approaching the State-of-the-Art (SoTA) Basque ASR baselines.

\textbf{Vocoder (Unit HiFi-GAN):} The vocoder synthesizes acoustic waveforms from discrete tokens. Its performance was measured using Mel Cepstral Distortion (MCD), where lower values (typically under 5-6 dB) indicate high synthesis quality. After training the model for 1000 epochs with 2 speakers, the Antton voice achieved a MCD score of 4.8~dB, and the Maider voice a score of 5.1~dB.

\subsection{End-to-End SLM}
We evaluated the full SLM pipeline using three metrics; the content quality of the text, the audio quality and the intelligibility of the speech response.

For content quality, we used an LLM-as-a-judge approach, prompting the Latxa-3.1-70B-Instruct model with a strict rubric to rate responses on a scale from 1 to 5 (LLM-as-a-judge Score) based on helpfulness, relevance, factual accuracy, and Basque fluency. This judge shares its family with our 8B backbone and 70B translation model (a possible self-preference bias); lacking a second judge or human annotators for Basque, we report this as a limitation. For audio quality, we used the UTMOS~\cite{saeki22c_interspeech} metric, which complements the vocoder's MCD reconstruction quality. This predictor was trained mainly on English and Japanese speech, so Basque scores are indicative rather than absolute. To assess speech intelligibility, we transcribed the generated audio using the state-of-the-art Basque Conformer ASR model\footnote{\href{https://huggingface.co/HiTZ/stt_eu_conformer_transducer_large_v2}{Link to the state-of-the-art Conformer model}} and calculated the WER and CER exclusively on Basque. All end-to-end scores are computed over a set of 4,419 synthetic instructions, disjoint from SLM training data; as both share the same synthetic pipeline and no real Basque speech was used at test time, this evaluation is in-distribution, and results on real user speech remain to be tested.

Our model achieves an average LLM-as-a-judge score of 3.86 out of 5. Regarding the spoken response evaluations, the model achieves a 3.07 UTMOS score, and an average WER of 9.94\% and a CER of 1.62\%.

When placed in the context of leading English multi-modal models (see Table \ref{tab:comparison_sota}), our model performs competitively. Despite Basque being a low-resource language, its LLM-as-a-judge score of 3.86 places closely behind LLaMA-Omni (3.99) and outperforms Qwen2-Audio (3.47) and SALMONN (3.44). Our model's ASR-WER of 9.94\% is comparable to LLaMA-Omni's English WER (10.82\%), with a remaining gap to LLaMA-Omni2-7B (3.26\%). The largest gap is in response speech quality, where our 3.07 UTMOS falls behind all English baselines. The streaming latency of 370 ms, measured on an NVIDIA A100 80GB GPU with chunk size $\Omega=40$ (matching LLaMA-Omni's setting), is close to LLaMA-Omni (346 ms) and far lower than cascaded pipelines like SpeechGPT (5587 ms).

\section{Conclusions}\label{section:6}

In this work, we developed the first open-source end-to-end Spoken Language Model (SLM) for Basque, establishing a reproducible framework to bridge the multi-modal gap for low-resource languages without massive foundational pre-training.

Our findings show that in the absence of spoken-instruction data, synthesizing data via an automated workflow is highly effective. Translating high-quality English corpora using LLMs combined with multi-speaker TTS synthesis provides a robust, scalable baseline for multi-modal alignment. The resulting model achieves a high content quality score of 3.86 out of 5, proving highly competitive with established English-centric systems.

Furthermore, our results show that attaching specialized speech adaptors and parallel streaming decoders to a language-specific text backbone successfully transfers native linguistic capabilities to the spoken modality. Architecturally, the model achieves excellent speech intelligibility with a Word Error Rate (WER) of 9.94\% for Basque words, comparable to that reported for the original English LLaMA-Omni across different languages and ASR systems, while maintaining a remarkably low streaming latency of 370 ms.

Future work will focus on integrating more expressive TTS systems to close the acoustic quality gap, complementing our automatic evaluation with a second judge model and a small-scale human evaluation, and testing the system on real, non-synthetic Basque speech.

\ifcameraready
\section{Acknowledgments}

This work has been partially funded by the projects 2022/TL22/00215335 ILENIA, and 'Desarrollo de modelos ALIA', funded by the Ministerio de Transformación Digital and by the Plan de Recuperación, Transformación y Resiliencia – Funded by the European Union – NextGenerationEU and by the project IkerGaitu II funded by the Basque Government. These models were trained at Hyperion, hosted by the DIPC Supercomputing
Center, and also at the Basque Government (EJIE) Data Processing Center.
\fi
\bibliographystyle{IEEEtran}
\bibliography{mybib}

\end{document}